\documentclass{llncs-modified}
\usepackage[T1]{fontenc}
\usepackage{graphicx}
\usepackage{booktabs}

\usepackage{hyperref}
\usepackage{color}

\begin{document}
\title{Optical Music Recognition for Real-World Manuscripts with Synthetic Data}
%
%
\author{Jiří~Mayer\inst{1,*} \and
 Martina~Dvořáková\inst{2} \and
 Vojtěch~Dvořák\inst{1} \and
 Markéta~Herzánová~Vlková\inst{2} \and
 Filip~Bím\inst{2} \and
 Pavel~Pecina\inst{1} \and
 Samuel Šomorjai\inst{2} \and
 Petr Žabička\inst{2} \and
 Jan Hajič jr.\inst{1}}
%
%
\institute{Institute of Formal and Applied Linguistics, Charles University \\
\and Moravian Library
\\ \email{* mayer@ufal.mff.cuni.cz}
}
%
\maketitle              
\begin{abstract}
Optical Music Recognition (OMR) has seen major progress in model design, with end-to-end methods now capable of recognising notation at all levels of complexity. However, the impact of this progress has been limited by the visual domains of available training datasets, which are largely born-digital. Existing large collections of sheet music in libraries and other heritage institutions contain predominantly manuscripts, whose visual domains are highly diverse and different, so existing OMR systems fail when applied in the real world. These institutions are often resource-constrained, so large in-domain datasets cannot be expected. We provide a first baseline on real-world manuscripts with complex piano notation in the resource-constrained scenario. Using fine-grained music notation graph (MuNG) annotations and the Smashcima synthesis tool, we then show that while some direct transcriptions of in-domain data remain essential, domain adaptation using synthetic musical manuscript images brings significant improvement. Furthermore, the symbols used do not need to be in-domain, so the expensive fine-grained annotation can be avoided. We thus bring OMR closer to one of its stated goals: preserving and promoting musical cultural heritage.

\end{abstract}

\section{Introduction}
\label{sec:introduction}


Optical Music Recognition (OMR), the field that aims to automatically read music notation, has been a particularly difficult sub-field of document recognition \cite{CalvoZaragoza-2020-understanding}. The difficulties related to the non-sequential and non-local nature of music notation have recently been largely overcome with attention mechanisms \cite{riosvila2024,Mayer2023Olimpic,RiosVila2025layout,RiosVila2024smtplusplus}. However, as these approaches require a substantial amount of supervised training data, a more insidious challenge remains: applying these models to real collections in resource-constrained setting.

\begin{figure}[t]
    \centering
    \includegraphics[width=1.0\linewidth]{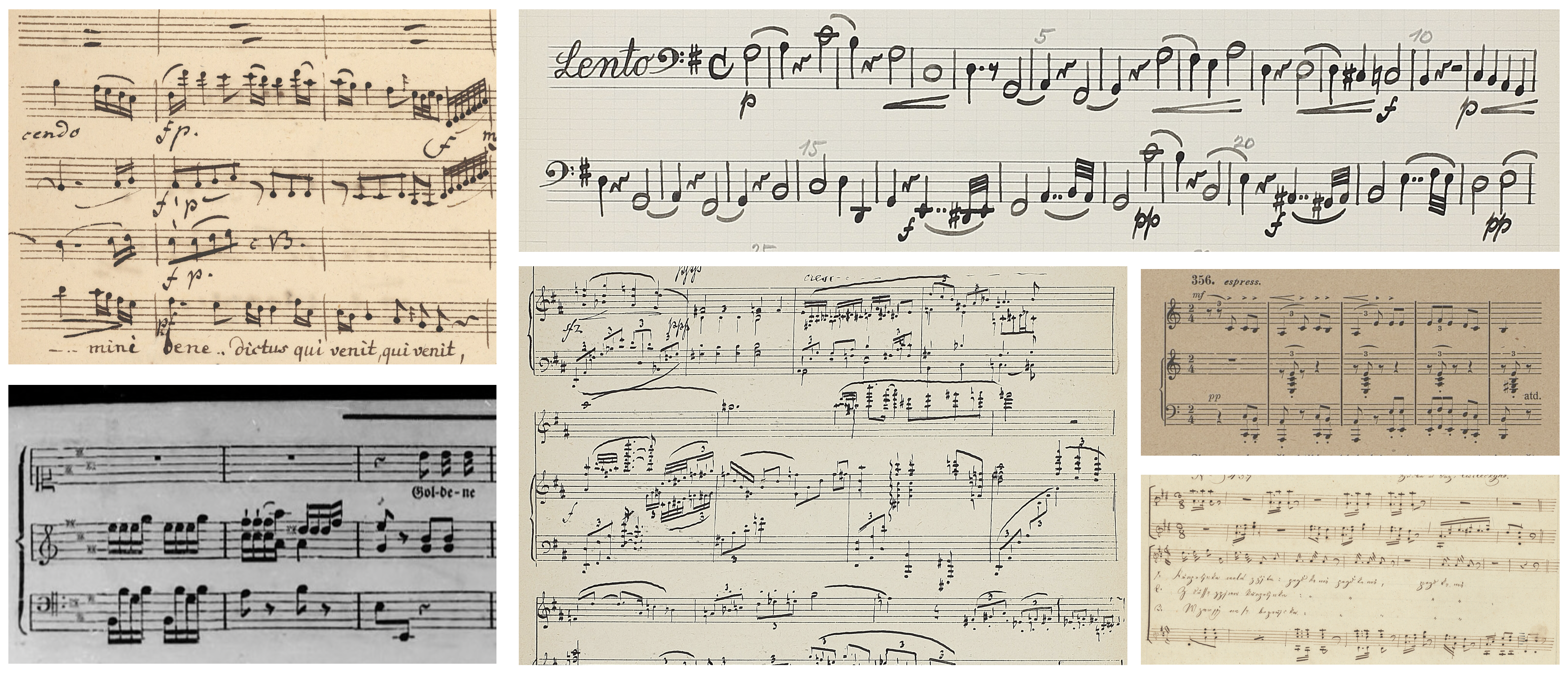}
    \caption{Examples of Common Western Music Notation in a library collection.}
    \label{fig:notation-diversity}
\end{figure}

Music notation documents are highly diverse (see Fig.~\ref{fig:notation-diversity}). 
As tools for digital production of new music notation are becoming increasingly  more and more integrated in musical workflows, the most likely large-scale users of OMR are increasingly music libraries and memory institutions, and perhaps music education institutions, rather than individual end users. The role of OMR in cultural heritage preservation, access and study thus requires systems to be able to process the collections of such institutions. The collections of music notation documents held in these institutions 
predominantly consist of manuscripts: the RISM database of musical sources in memory institutions records 1,346,175 manuscripts, compared to 261,162 prints.\footnote{As of February 6, 2026: \url{https://rism.online/}} To provide, e.g., retrieval services \cite{diet2018staatsbibliothek,hajic2018useful,crawford2023exploring,umbreit2024omr}, OMR therefore needs to work across a variety of visual domains, including handwritten music. Domain adaptation has recently been identified in surveys as a key challenge for OMR \cite{CalvoZaragoza2023review,Castellanos2025review}.

However, OMR currently has no dataset of real-world musical manuscripts publicly available. And because producing in-domain ground truth for training OMR on musical manuscripts is an expensive endeavour \cite{Torras2025TwoJourneys}, it is unlikely that building sufficiently large in-domain datasets will be within the means of user institutions any time soon. 

An obvious next-best solution in the absence of ``real'' data is synthetic data. A system for OMR data synthesis needs to output two components necessary for supervised learning: (1) the musical content that should be available in notation and (2) the image that corresponds to the music notation encoding this musical content.
The first component presents few issues. Public-domain datasets of machine-readable representations of diverse music that can serve as targets for end-to-end learning, in symbolic formats such as MusicXML \cite{GothamJonas2022,Long2025PDMX}, ABC \cite{wang2025notagen}, or \texttt{**kern},\footnote{\url{https://kern.ccarh.org/}} are available. 
However, the problem is on the side of actually producing score images.
Many tools for rendering music notation from its symbolic encoding exist: regular notation editors such as MuseScore\footnote{\url{https://musescore.com/}} (which has a batch processing mode), Sibelius\footnote{\url{https://www.avid.com/sibelius}} or Dorico,\footnote{\url{https://www.steinberg.net/dorico/}} the TeX-based LilyPond system, or the Verovio renderer for the MEI format. However, all of these (naturally) only render print-like notation that furthermore does not even entirely generalize to scans of music prints, even with data augmentation \cite{Mayer2023Olimpic}. A certain ``glass ceiling'' has been observed \cite{AlfaroContreras2023}.

None of this notation software is capable of rendering images of musical manuscripts. Attempts have been made in style transfer from print to manuscript \cite{Shatri2024GANs,Pranjali2023,Tirupati2024}, for a workflow where the manuscript image is obtained as a purely visual transformation of a rendered print image while not ``damaging'' the musical semantics, but these have not led to a workable solution so far.

The remaining pathway is therefore to implement a music notation rendering engine that could render correct manuscript images directly from symbolic representations. Previous experiments with limited synthetic manuscript data have shown some promise \cite{Mayer2021Mashcima,Mayer2022Obstacles}. However, these systems have been limited to very simple music and still unrealistic images, and this ``promise'' has never been confronted with real-world scores in realistic settings.


Recently, the Smashcima system has been introduced \cite{Mayer2025Smashcima} that is capable of rendering full-page manuscript images from (nearly) arbitrary MusicXML inputs. It extracts glyph libraries from the Music Notation Graph (MuNG) representation, introduced in the MUSCIMA++ dataset \cite{MUSCIMAPP}. 
In combination with the MuNG Studio tool for efficiently annotating sheet music in this representation \cite{Mayer2025MuNGStudio}, the infrastructure is now in place to leverage manuscript synthesis for domain adaptation to musical manuscripts, even for diverse collections of resource-constrained memory institutions. 

However, experimental work to show whether this pathway is in fact useful is still missing.

\begin{figure}[t]
    \centering
    \includegraphics[width=1.0\linewidth]{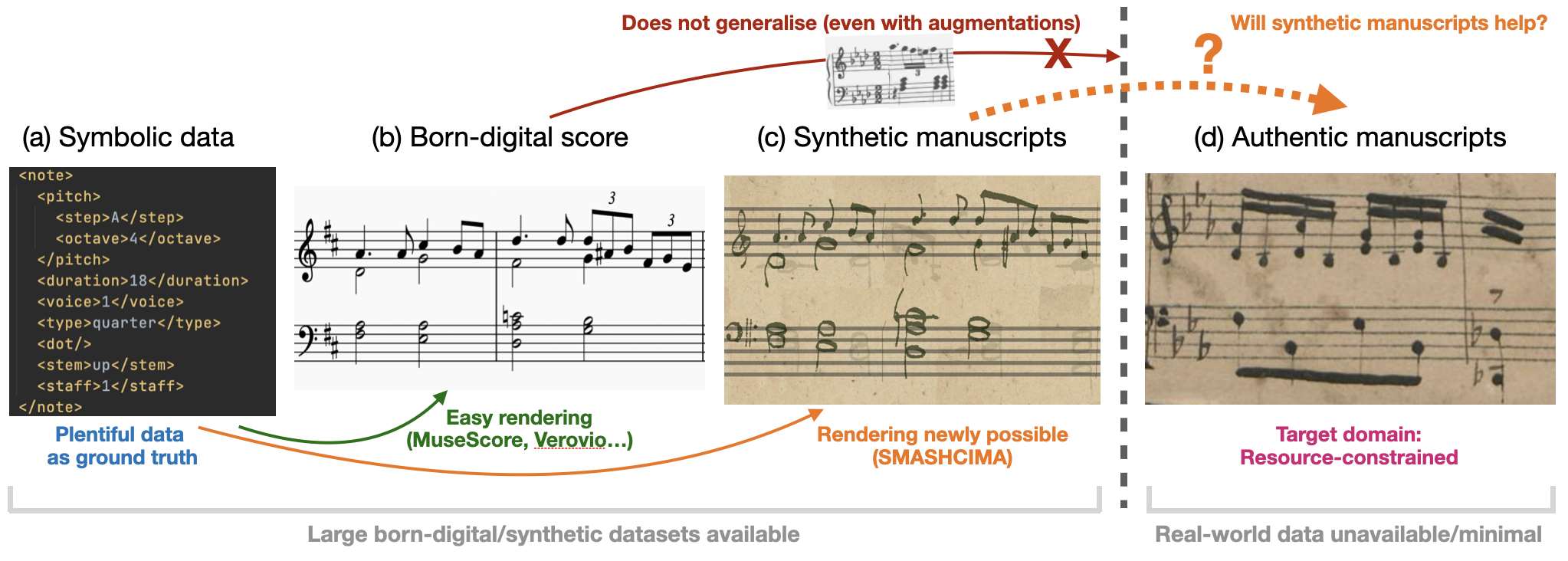}
    \caption{In this paper we experiment with newly available synthetic manuscript images of sheet music to see whether they help with domain adaptation to authentic (real) manuscripts.}
    \label{fig:paper-overview}
\end{figure}

As indicated in Fig.~\ref{fig:paper-overview}, this is what we contribute. Simulating the scenario of large available symbolic datasets and very limited visually diverse in-domain data (primarily manuscripts) from the musical collections of a resource-constrained library,
we find that realistic synthetic manuscript data lead to useful improvement in domain adaptation on real-world diverse manuscripts.

After reviewing related work in Sec.~\ref{sec:relatedwork}, 
we describe the data synthesis setup in Sec.~\ref{sec:synthesis}, the datasets used in Sec.~\ref{sec:datasets},
and the experimental setup in Sec.~\ref{sec:experiments}.
Sec.~\ref{sec:results} describes experiment results and Sec.~\ref{sec:conclusions} discusses the implications of the experimental results for OMR and cultural heritage preservation, 
limitations of the current setup, sketch out future work, 
and highlights the interaction between generic machine learning methods and adequate domain-specific software infrastructure that makes this possible.

\section{Related Work}
\label{sec:relatedwork}

An essential context for our work are the successes of attention-based sequence-to-sequence learning methods in dealing with the complexities of Common Western Music Notation (CWMN) \cite{riosvila2024,Mayer2023Olimpic}, even at the full page level \cite{RiosVila2024smtplusplus}. 
(Vision large language models have so far not been able to contribute to OMR beyond replicating state-of-the-art results on a born-digital synthetic dataset \cite{calvo2024llms,steiner2024paligemma2familyversatile}.)
OMR models are thus no longer limited by what is in the music --- rather, they are limited by what the target music looks like.

Despite manuscripts making up the majority of CWMN collections, however, none of these experiments were performed with manuscript data. So far, beyond the first attempts on CWMN manuscripts based on the object detection pipeline \cite{Hajicjr.2018a,Pacha2019Assembly}, there has been just one attempt with domain adaptation and BLSTMs \cite{Baro2019HmrBaseline}, and these works were limited to the MUSCIMA++ dataset, which is binarized and contains notation written in the 21st century \cite{CVC-MUSCIMA,MUSCIMAPP}. Therefore, how the advance in OMR models maps to real-world conditions in music libraries is unknown. We believe this foregrounds domain adaptation as a leading current challenge for OMR.

This is fundamentally connected to data availability: there is no open dataset of  CWMN manuscript images available for training OMR systems, with the small exception of the monophonic 20th-century FMT \cite{JuanCarlos2024universal}.
Hence, no study on the impact of CWMN synthetic data on real-world OMR has been performed.

However, the potential of data synthesis for a resource-poor field such as OMR, especially on handwritten notation, has been recognised for some time. First attempts at synthesizing handwritten music images come from Baró et al. \cite{Baro2019HmrBaseline} with their approach of measure shuffling the MUSCIMA++ dataset. The logical extension of this idea to individual symbols was performed by Mayer et al. with the Mashcima system \cite{Mayer2021Mashcima}. Many of its limitations (monophonic music, non-standard encoding, no customisability) were recently overcome with the successor Smashcima system \cite{Mayer2025Smashcima}.

The avenue of deep-learning synthesis approaches has been explored by many different people. The team around Elona Shatri has done experiments with generative adversarial networks \cite{Shatri2024GANs,Pranjali2023,Tirupati2024}. Autoencoders were used to synthesize individual symbols \cite{Havelka2023}. Recently, researchers from the MALer lab have built a system that can produce synthetic images of music notation through an RQVAE decoder \cite{UmusTKorea2025}.

Manuscript data for earlier European notations (medieval and renaissance-period) is available, and experiments across four different datasets of mensural notation, of which two were handwritten, indicated that synthetic data may indeed help domain adaptation \cite{JuanCarlos2023}. Experiments on different medieval Gregorian chant datasets show that different real-world visual domains may actually be best addressed with somewhat different model architectures \cite{FuentesMartinez2026}; they also highlight the potential of synthetic data as part of pre-training, but domain adaptation has not yet been studied there.

Outside of OMR, the widespread use of synthetic data for manuscripts has been reviewed in 2024 \cite{deSousaNeto2024review}, noting that image generation models were increasingly being used to generate synthetic data for low-resource settings. Content and style conditioned synthesis was achieved in handwriting \cite{PauRibaGANWriting}, but it still remains elusive in the music domain.






\section{OMR data synthesis process}
\label{sec:synthesis}


As with any data for supervised learning, a synthesis engine for OMR works in the opposite direction of an OMR system: it produces a sheet music image correctly reflecting a given ground truth encoding of music notation, such as MusicXML. The synthesiser essentially engraves the given music.

Due to the nature of music notation as a writing system \cite{CalvoZaragoza-2020-understanding}, this is a very complex task.
Music notation editors such as MuseScore, Sibelius, or Verovio do this;
however, their outputs are restricted to their own fonts, which naturally prioritise a ``clean'', consistent look with just one shape per glyph type, certainly not simulating manuscripts with their inherent variability.
As reviewed in Sec.~\ref{sec:relatedwork}, efforts to train style transfer-like models that would then deform the resulting image to look like a manuscript without loss of musical semantics have so far not been successful. 

The other pathway is to build a music notation engraving engine that would be capable of rendering simulated manuscript images directly.
Fortunately, the recently released Smashcima tool \cite{Mayer2025Smashcima} acts like a music notation rendering engine: it can render a MusicXML file using large symbol sets.
This finally opens an opportunity to perform visual domain adaptation without being restricted to oversimplified notation styles (such as in \cite{Mayer2021Mashcima} or \cite{Baro2019HmrBaseline}).

\begin{figure}
    \centering
    \includegraphics[width=0.8\linewidth]{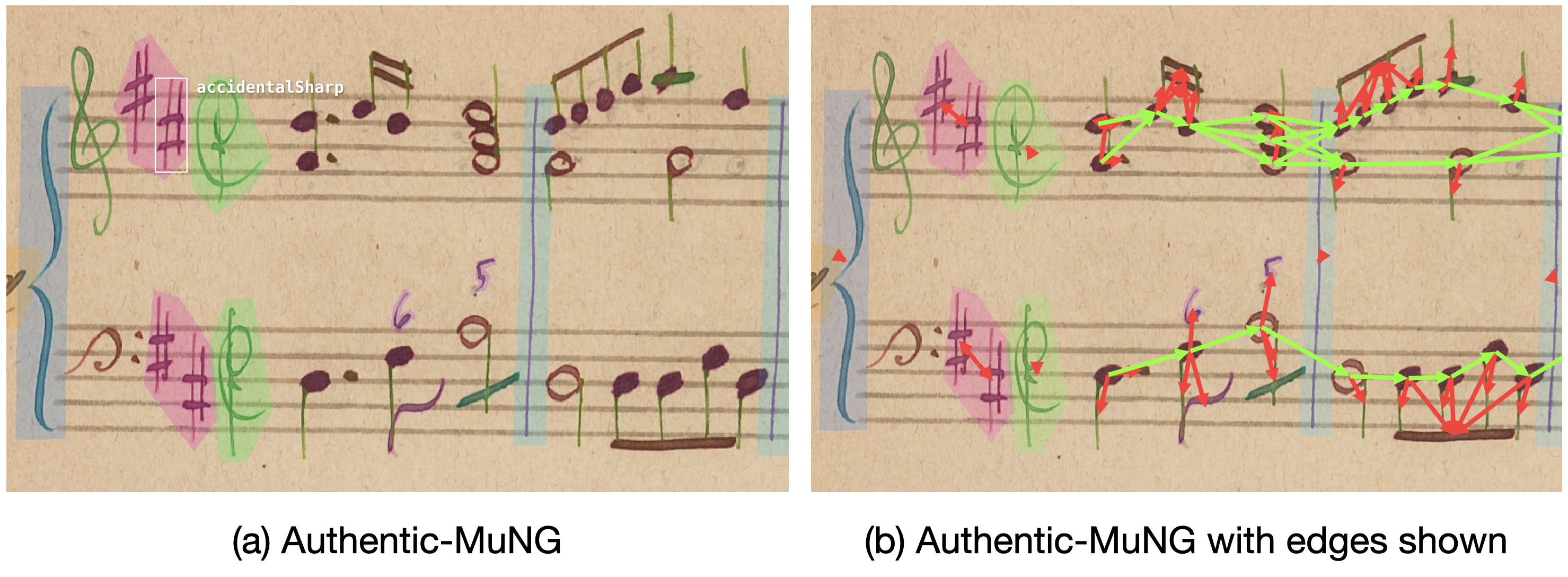}
    \caption{Example annotation in the MuNG format. On the left, just the symbols are shown; note the accuracy of the symbol masks (e.g., the G-clef or the 8th flag in the first measure of the bottom staff). On the right, the annotation is shown with edges.}
    \label{fig:mung-example}
\end{figure}

\begin{figure}[b]
    \centering
    \includegraphics[width=1.0\linewidth]{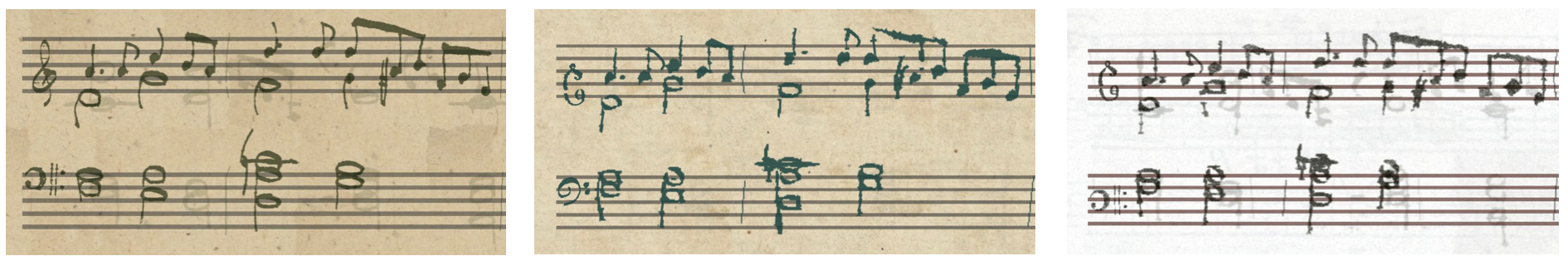}
    \caption{Different renderings of the same MusicXML file using Smashcima.}
    \label{fig:smashcima-outputs}
\end{figure}

Smashcima requires three inputs to operate.
Most crucially, it needs symbol sets from the target domain. 
These are supplied in the MuNG format, established in the MUSCIMA++ dataset \cite{MUSCIMAPP}. An example MuNG annotation is shown in Fig.~\ref{fig:mung-example}. Annotating music documents in the MuNG format is cca 10x as expensive as transcribing them in MusicXML \cite{Torras2025TwoJourneys}.
%
The second input Smashcima needs are several samples of expected backgrounds, which can easily be taken from in-domain documents.
The third input are MusicXML files encoding the sheet music to be rendered. 
These are plentiful, e.g., the ground truth of datasets with born-digital images can be used.
Smashcima can also apply a number of standard document image augmentation techniques as postprocessing steps (noise, ink simulation, bleedthrough, deformations, scribbles, shadows and other camera effects, etc.).
Example outputs of synthesising the same MusicXML in different styles with Smashcima
are shown in Fig.~\ref{fig:smashcima-outputs}.

\subsection{Perception study}

In a small user study, we have found that images rendered by Smashcima are barely recognisable for humans from authentic scores. We ran a survey with 10 authentic and 10 synthetic images, asking for each image whether it was authentic, or synthesised. The survey was shuffled randomly for each participant to avoid priming effects. A total of 30 participants marked each presented image as synthetic or authentic, leading to 600 binary responses in total, out of which 300 were on synthetic examples. While participants still could recognise synthetic images at higher than chance levels (posterior CDF for H$_0$: $P(synth.guessed \mid synth.true) \leq 0.5$ was $<0.001$ under the Beta-Bernoulli model with prior $\alpha, \beta = 1$), their overall accuracy was merely $0.61$ (effect size $0.11$), and participants who were able to read music notation (23 out of the 30) were in fact worse, with overall accuracy $0.57$ (effect size $0.07$, but still above chance level, with posterior CDF for H$_0$ at $0.013$).

\section{Datasets}
\label{sec:datasets}

The composition of datasets used in this experiments reflects the ``resource-constrained library'' scenario outlined above: large out-of-domain datasets with born-digital images, and very small in-domain (``authentic'') datasets. The one exception is in-domain data in MuNG format for synthesis, which would require a significant resource investment: in the spirit of providing practical recommendations, we attempt to find out whether this extra investment (rather than just transcribing to MusicXML) leads to significantly better results than using non-authentic (out-of-domain) MuNG data.

To ensure the complexity of real music notation is accounted for, and to facilitate comparability with the most recent end-to-end OMR experiments, the setup focuses on piano notation \cite{CalvoZaragoza-2020-understanding}. We also operate on single-system\footnote{A system in music notation means one line of music of all instruments or voices that are playing simultaneously; typically this is indicated by a brace on the left side of the system, and vertical measure-separating lines running across all staves within one system. In solo piano music, a system consists of music on two staffs: one for the right hand, one for the left.} data as most of OMR has: recently, it has been shown that end-to-end models are capable of generalising from single-staff to full-page sheet music images using curriculum learning \cite{RiosVila2024smtplusplus}, so the distinction between single-system and full-page settings is not as relevant to the domain adaptation setting: if synthetic data lead to large improvements for end-to-end models on single-staff images, then the same method should generalise these results to full pages.

In order to measure how synthetic data help in domain adaptation, we need three types of datasets: 
\begin{itemize}
    \item Large out-of-domain datasets, representing the plentiful born-digital data,
    \item Small in-domain (authentic) datasets, representing outputs of the resource-constrained annotation: cheaper end-to-end ground truth (MusicXML) for fine-tuning, and possibly expensive data (MuNG) for data synthesis.
    \item Large synthetic datasets, created by combining the out-of-domain ground truth with in-domain symbol data.
\end{itemize}

\subsection{Out-of-domain Datasets}

We use two existing datasets of the first type: GrandStaff \cite{Rios-Vila-etal-2023-end-to-end} and OLiMPiC \cite{Mayer2023Olimpic}. Both datasets define default train/test splits that we follow. 

The \textbf{GrandStaff} dataset \cite{Rios-Vila-etal-2023-end-to-end} is a subset of piano music from the KernScores database\footnote{\url{https://kern.ccarh.org/}} that contains scores of 474 compositions by 6 composers. All compositions are transposed into 3 additional keys, segmented into chunks of 3-6 measures, and each of these chunks was rendered by Verovio \cite{Pugin_2014} as a JPEG image. Each chunk serves as one data point, for a total dataset size of 53,882 samples.
The \textbf{Camera-GrandStaff} dataset is exactly the same as GrandStaff, except that each rendered JPEG image is further distorted \cite{Calvo-Zaragoza2018CameraPrimus}. 
In order to make GrandStaff ground truth compatible with other datasets, we use its version encoded in LMX (Linearised MusicXML \cite{Mayer2023Olimpic}).

The \textbf{OLiMPiC} dataset \cite{Mayer2023Olimpic} is derived from OpenScore Lieder \cite{OpenScore.2018,GothamJonas2022}, a corpus of music encodings. The dataset consists of piano accompaniments for 1,295 different French and German 19th-century songs by 107 distinct composers, also cut into systems for a total of 17,945 data points consisting of a rendered image of a system (using MuseScore 3.6.2).\footnote{This is in fact a more musically diverse dataset than GrandStaff: without transpositions, GrandStaff has 13,470 distinct lines of music.}
As with GrandStaff, we encode the MusicXML ground truth in LMX.
%

\begin{figure}[t]
    \centering
    \includegraphics[width=0.8\linewidth]{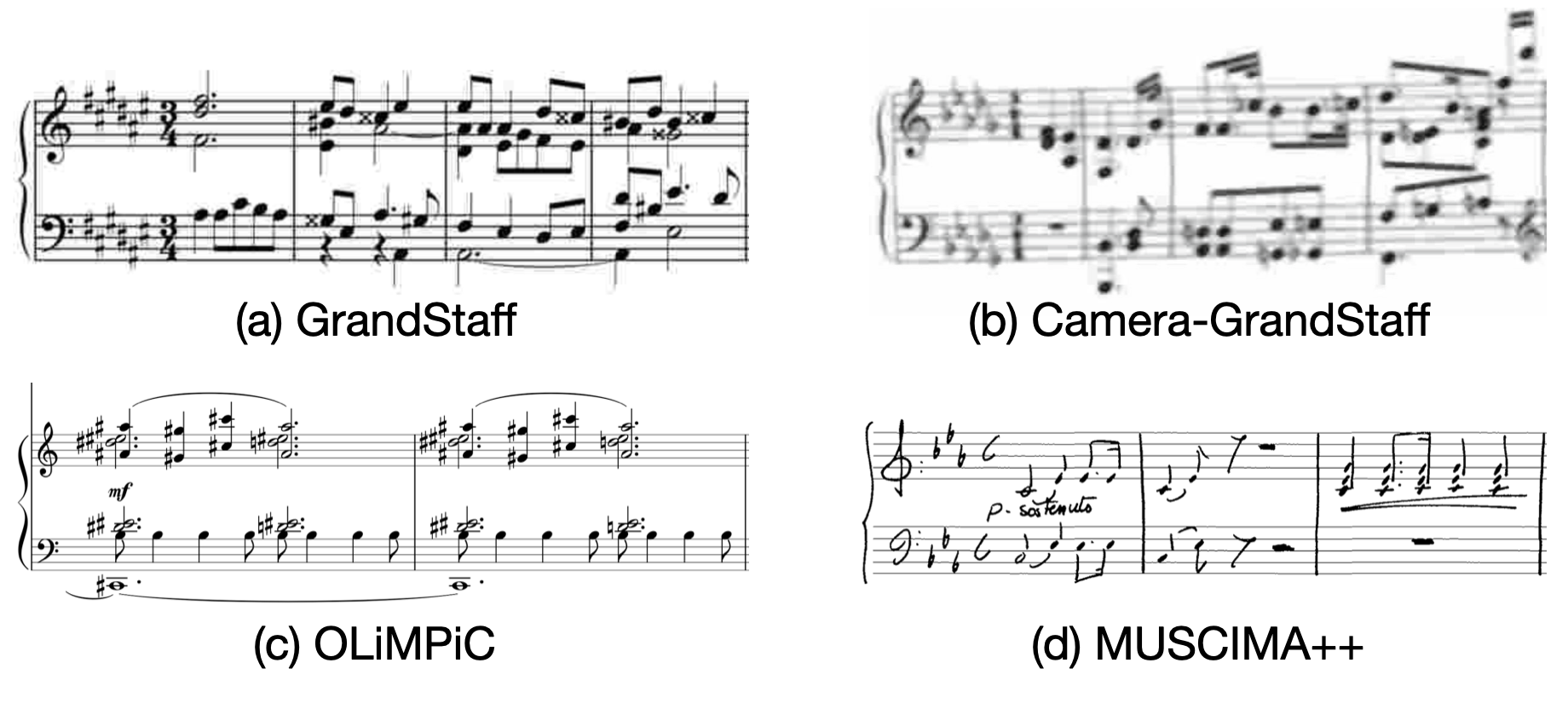}
    \caption{Examples of the variety of out-of-domain datasets.}
    \label{fig:out-of-domain-examples}
\end{figure}

The \textbf{MUSICMA++} dataset \cite{MUSCIMAPP} serves as an alternative source of handwritten symbols, out-of-domain with respect to the library collection, for creating synthetic data with Smashcima. It consists of 91,255 annotated notation symbols. In this work, it is used to evaluate the importance of using in-domain symbols as inputs for the data synthesis process, compared to generic handwritten symbols.
Samples from these out-of-domain datasets are shown in Figure~\ref{fig:out-of-domain-examples}.

\subsection{In-domain Datasets}


The \textbf{Authentic} dataset is at the heart of the experiment setup: it represents the small in-domain dataset for end-to-end recognition used for fine-tuning experiments and, most importantly, evaluating the quality of domain adaptation.
It comes from the collections of the Moravian Library, as a representative sample selected by its music librarians. It consists of 159 piano systems that have been transcribed in MuseScore, exported to MusicXML and encoded in LMX. A 59:50:50 train-validation-test split is defined by assigning the systems randomly. 
The total effort to transcribe this dataset was 25 hours.

\begin{figure}[t]
    \centering
    \includegraphics[width=1.0\linewidth]{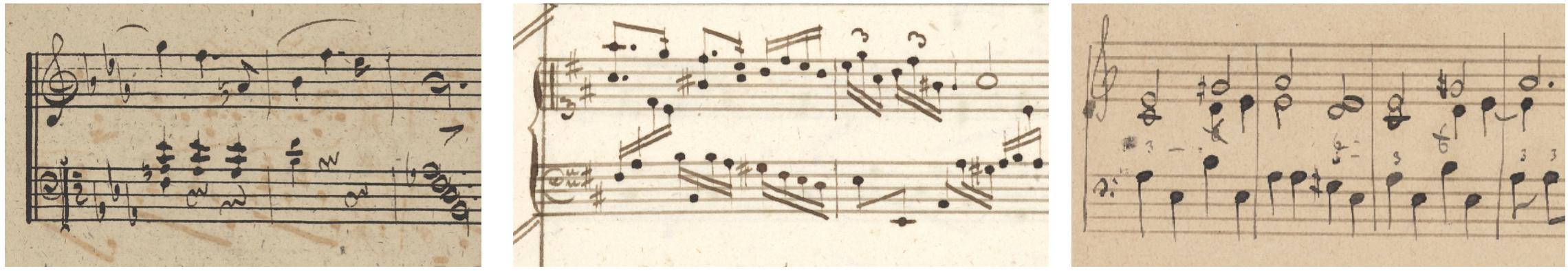}
    \caption{Examples of images from the Authentic dataset.}
    \label{fig:authentic-examples}
\end{figure}

The \textbf{Authentic-MuNG} dataset serves as the source of input symbols for creating synthetic data. For simplicity, we exploit the same 59 samples from the training (or fine-tuning) portion of the Authentic dataset, but annotated in the MuNG format. The total number of annotated symbols across all splits is 39,376 but only 15,011 are in the training split, which is used for synthesis. The total effort to annotate this dataset was approximately 260 hours.

\begin{table}[t]
    \centering
    \caption{List of datasets split into sections according to the description above. Datasets with bold names are those that had to be created by the library in our scenario. The ``Function'' column describes how the dataset is used in our experiments: ``synthesis'' means that the dataset is used as input to Smashcima and its ground truth are symbols annotated as MuNG rather than LMX-encoded music transcriptions.}
    \begin{tabular}{l@{\hskip 3pt}|@{\hskip 3pt}lllrl}
         \textbf{Name} & \textbf{Domain} & \textbf{Production} & \textbf{Function} & \textbf{Size} & \textbf{Unit} \\
         \hline
         GrandStaff & out & born-digital & training & 53,882 & systems \\
         Camera-GrandStaff & out & born-digital & training & 53,882 & systems \\
         OLiMPiC & out & born-digital & training & 17,945 & systems \\
         \hline
         \textbf{Authentic} & in & manuscript & eval+fine-tuning & 159 & systems \\
         \textbf{Authentic-MuNG} & in & manuscript & synthesis & 39,376 & symbols \\ 
         MUSCIMA++ & out & manuscript & synthesis & 91,255 & symbols \\
         \hline
         Synth-GS-Auth & in & synthetic & training & 44,192 & systems \\
         Synth-OL-Auth & in & synthetic & training & 16,273 & systems \\
         Synth-GS-MPP & out & synthetic & training & 44,191 & systems \\
         Synth-OL-MPP & out & synthetic & training & 16,273 & systems \\
    \end{tabular}
    \label{tab:datasets}
\end{table}

\subsection{Synthetic Datasets}

Finally, we describe the datasets synthesised using the Smashcima tool. As described in Sec.~\ref{sec:synthesis}, it requires three inputs:

\begin{itemize}
    \item MusicXML -- musical content; taken from the GrandStaff and OLiMPiC datasets.
    \item Music notation glyphs -- capture the visual domain of a document; taken from MUSCIMA++ and the Authentic-MuNG datasets. 
    \item Background patches -- we use 13 patches from the MZKBlank set \cite{Dvorak2024}.
\end{itemize}

The \textbf{Synth-GS-Auth} dataset is created by running the Smashcima system with MusicXML inputs from the GrandStaff dataset and using symbol glyphs from the \textbf{Authentic-MuNG} dataset. The \textbf{Synth-GS-MPP} dataset is the same, but uses symbols from MUSCIMA++ instead. Similarly \textbf{Synth-OL-Auth} and \textbf{Synth-OL-MPP} are constructed in the same manner from MusicXML data from the OLiMPiC dataset. The two datasets that use MUSCIMA++ symbols are not in-domain since they do not use Authentic symbols but their purpose is to estimate the benefit of having in-domain symbols in the synthetic data. When running Smashcima, we lose 9~\% of OLiMPiC samples and 18~\% of GrandStaff samples due to incompatibilities and crashes (e.g., unsupported rare symbol classes; some of these come up more often in GrandStaff because of its transpositions).

\section{Experiments}
\label{sec:experiments}

Each experiment consists of three phases. \textbf{First,} we take a randomly-initialized model and train it on a base dataset (Camera-GrandStaff, OLiMPiC, or one of the synthetic datasets). This training takes the longest amount of time (cca 3 days on a A40 GPU). 
\textbf{Second,} we ``re-train'' this base model, continuing the training on synthetic data (only for non-synthetic base datasets). This phase is meant to explore adaptation of existing pre-trained OMR models. This re-training phase is shorter, but still takes around 2 to 8 hours. 
\textbf{Third,} we fine-tune on the Authentic dataset. This is the fastest phase (cca 10 minutes) because the Authentic dataset is small and prone to overfitting.

In the second and third phase we do not use any data from the previous phase (that is, we do not re-train with replay). This way the model risks catastrophic forgetting of the previously learned data. As our aim is to adapt to the target domain, not to build a general-purpose model, this does not matter. In the last experiment, however, we perform re-training \textit{with} replay\footnote{Mixing the synthetic images with the born-digital images.} to see what effect it has on the final result.

The initial training phase uses learning rate of $1e^{-3}$ with cosine decay over 150 or 500 epochs on GrandStaff or OLiMPiC respectively (GrandStaff is about 3x the size of OLiMPiC). The two re-training and fine-tuning phases use constant learning rate of $1e^{-4}$.

The re-training phase stores a model snapshot after each epoch and then we find the best epoch. This turns out to be very short without replay (2-10 epochs) due to the forgetting-learning tension; with replay, the re-training can proceed for 30 epochs with no issues.

The final fine-tuning phase always takes 20 epochs, which we estimated by observing the symbol error rate (SER) stopping decreasing on the validation split of the Authentic dataset. This is stable regardless of which model is being fine-tuned.

\subsection{Model}

We use the Zeus model, which holds (together with a fine-tuned PaliGemma 2) the reported state-of-the-art SER on the GrandStaff dataset \cite{Mayer2023Olimpic}. The only relevant difference between Zeus and the Sheet Music Transformer (SMT) \cite{riosvila2024} is the use of soft attention with recurrent layers rather than Transformer self-attention and positional encodings, and Zeus led to a nearly 50 \% relative error reduction on the GrandStaff dataset \cite{Mayer2023Olimpic}, we use this model. However, in principle, one could also use SMT; the experimental conditions are not different models but different dataset choices.
%
%
The exact settings with which the model is run are the same as in \cite{Mayer2023Olimpic}.

\subsection{Evaluation}

We perform all evaluation on the test split of the Authentic dataset, as it represents the unseen manuscripts from the library we want to adapt to. We use the Symbol Error Rate (SER), a metric reported in most end-to-end OMR works \cite{calvo-zaragoza-etal-2018-end-to-end,RiosVila2022,riosvila2024,Mayer2023Olimpic,AlfaroContreras2023,RiosVila2024smtplusplus,Jung2025uMust}. Evaluation of OMR is a notoriously hard problem \cite{Byrd2015,Hajicjr.2016,torrasPauMTN} but SER has been shown to correlate well with suggested improvements \cite{torrasPauMTN,JuanCarlos2025benchmark}. Previous in-domain results on GrandStaff and OLiMPiC reported at a SER around 2.0 and 11.0 respectively \cite{Mayer2023Olimpic,steiner2024paligemma2familyversatile,Jung2025uMust,riosvila2024}.

The test split of the Authentic dataset has 50 systems of music. This is of course a small sample with high variance, however, we are interested in qualitative leaps, not in competing for the final few percentage points, and we believe the variance in this case is acceptable; in any case, no larger real-world dataset of musical manuscripts is currently available. Cross-validating the best-performing model across all 6 permutations of the fine-tuning, validation, and test portion of the Authentic dataset, we find that the standard deviation across the Authentic dataset is $4.2$, so a difference of some 8.4 SER on this dataset is 95\% likely meaningful (and a difference of 4.2 SER is 68\% likely meaningful).
%


\section{Results}
\label{sec:results}

We perform a total of 11 experiments grouped into four scenarios as indicated in Table \ref{tab:results}.
Experiments 1 and 2 act as baselines where we only fine-tune existing OMR models on our Authentic data: one trained on the Camera-GrandStaff dataset, the other on OLiMPiC. The far better base model is the Zeus OLiMPiC model from 2024: from a  SER of 88.0, when naively applied to our Authentic test set, it improves to 41.3 after fine-tuning (while the base model on Camera-GrandStaff fine-tunes to 85.2). The OLiMPiC base model result is stable within 1 SER when re-run with different seeds, thus forming the baseline for all future models that use synthetic data. If none of them significantly pass 41.3 SER, synthetic data brings no advantage compared to just fine-tuning.

Experiments 3 to 6 try building a base model on synthetic data only. We see that applying these models naively results in SER of 56.9--76.3, which is better than the 88.0 from Experiment 2; however, after fine-tuning this drops to only 51.4--58.5 SER. Though these models seem better suited for handwritten music, the Zeus OLiMPiC model proves more adaptable. What is, however, clear from pre- and post- fine-tuning evaluation is that Authentic symbols outmatch MUSCIMA++ symbols. This makes a case for using in-domain glyphs when building the synthetic dataset. 

Experiments 7 to 10 take the pre-trained Zeus OLiMPiC model and then train it further on our four synthetic datasets. This re-training happens without replay of OLiMPiC data, which causes a tension between forgetting of typeset music and learning of handwritten music. The optimal re-training epoch is between 2 and 6, based on the lowest SER (after fine-tuning) on the Authentic validation split. Evaluating the re-trained models shows a drop of cca 7.0 SER compared to Experiments 3 to 6 (before fine-tuning), and the preference for authentic glyphs is still significantly present. However, \textit{after} fine-tuning, the preference disappears: models improve to around 34.6 SER in all experiments. This is the first setup that managed to beat the 41.3 SER fine-tuning baseline (Ex 2).

To avoid the tension of Experiments 7 to 10, we designed Experiment 11 to include data replay. The re-training dataset is the OLiMPiC and Synth-OL-Auth datasets combined into one (leading to twice as many samples per epoch). This lets us not worry about early stopping (before forgetting important OLiMPiC traits) and we can now train for 30 epochs (equivalent to 60 epochs in Experiments 7 to 10). With this setup, we manage to just outperform the Ex 2 baseline without even fine-tuning, at 38.1 SER. If we now also fine-tune, we improve to 33.2 SER, which is technically the best result (though by little).

\begin{table}[t]
    \centering
    \caption{Main experimental results: For each experiment a new model is created, trained on the first dataset, then re-trained (without replay) on the second one, and finally fine-tuned (no replay) on the Authentic dataset (training split). Symbol error rate (SER) is reported after each step on the Authentic dataset (test split). Number of training epochs is shown in parentheses. Fine-tuning is done for 20 epochs. (SER >100 can happen when the model generates many extra tokens.)}
    \begin{tabular}{l|@{\hskip 3pt}lc@{\hskip 3pt}|@{\hskip 3pt}lc@{\hskip 3pt}|c}
         \toprule
         \textbf{Ex} & \textbf{Trained on} & \textbf{SER} & \textbf{Re-trained on} & \textbf{SER} & \textbf{SER ft}  \\
         \hline
         1 & Camera-GrandStaff {\tiny (150e)} & 122.9 & -- & -- & 85.2 \\
         2 & OLiMPiC {\tiny (500e)} & 88.0 & -- & -- & \textbf{41.3} \\
         \hline
         3 & Synth-GS-MPP {\tiny (150e)} & 70.9 & -- & -- & 56.3 \\
         4 & Synth-GS-Auth {\tiny (150e)} & 56.9 & -- & -- & 52.5 \\
         5 & Synth-OL-MPP {\tiny (500e)} & 76.3 & -- & -- & 58.5 \\
         6 & Synth-OL-Auth {\tiny (500e)} & 62.8 & -- & -- & 51.4 \\
         \hline
         7 & OLiMPiC {\tiny (500e)} & 88.0 & Synth-GS-MPP {\tiny (2e)} & 63.4 & 34.7 \\
         8 & OLiMPiC {\tiny (500e)} & 88.0 & Synth-GS-Auth {\tiny (2e)} & 51.8 & 34.6 \\
         9 & OLiMPiC {\tiny (500e)} & 88.0 & Synth-OL-MPP {\tiny (6e)} & 63.1 & 34.8 \\
         10 & OLiMPiC {\tiny (500e)} & 88.0 & Synth-OL-Auth {\tiny (2e)} & 49.9 & 34.4 \\
         \hline
         11 & OLiMPiC {\tiny (500e)} & 88.0 & Synth-OL-Auth w/ replay {\tiny (30e)} & \textbf{38.1} & \textbf{33.2} \\
         \bottomrule
    \end{tabular}
    \label{tab:results}
\end{table}

\section{Discussion and Conclusions}
\label{sec:conclusions}

For the first time, we have run OMR experiments with real-world manuscripts of piano music, the most complex class of Common Western Music Notation. We have thus established a baseline for this most difficult, but most impactful, setting for OMR: low-resource adaptation to diverse historical manuscripts. 

A SER of 33.2 is of course much higher than the previously reported results on simple datasets such as GrandStaff or OLiMPiC with SER approx. 2.0 and 11.0 respectively \cite{riosvila2024,Mayer2023Olimpic,steiner2024paligemma2familyversatile}. However, it must be stressed that these results were achieved for clean born-digital images, made more complex only by a limited set of augmentations. While these datasets are good benchmarks for model comparison, these results are not representative of expected performance in application settings at all. Already on the scanned version of OLiMPiC evaluation data, which are all high-quality scans of high-quality prints, SER deteriorates to 17.7 \cite{Mayer2023Olimpic}. In other words --- the OMR \textit{models} are fine, but the task of making OMR \textit{systems} valuable is just beginning.

To our main question: Are high-quality synthetic data in turn valuable for training OMR systems in difficult low-resource settings?

On the one hand, the answer is yes. The presence of synthetic data matched to the target domain did lead to significantly better results than when this class of data was absent. 
On the other hand, the results are certainly not the qualitative leap one may have imagined on the basis of impressions of the data synthesis results (Fig.~\ref{fig:smashcima-outputs}), and of the near-inability of people (incl. those capable of reading music) to distinguish Smashcima-generated from real images.

However, the experiments revealed a silver lining. After fine-tuning, the advantage of using in-domain MuNG data for Smashcima over using the MUSCIMA++ symbols disappears. Therefore, it seems that to get the available mileage out of Smashcima, one actually does not need to take on the expense of annotating in-domain MuNG data. 
One needs authentic fine-tuning data anyway, and the final system quality will likely correlate to the amount and diversity of this data. But adding a synthetic dataset into the training process does provide a significant boost, especially when fine-tuning data is scarce; at least, this implies that less fine-tuning data will be needed to achieve the same performance when Smashcima is used.

What seems as a viable recommendation for OMR system development is not to treat the synthetic data as part of domain adaptation, but to mix them directly into the base model training.

As evidenced by the impact of fine-tuning with even a very small authentic dataset, Smashcima is still not able to entirely simulate the domain of real manuscript music notation. The limitations mostly concern how symbols connect: we noticed, e.g., that stems only attach to noteheads on the rightmost point, while many historical writing styles attach on the top of the notehead. However, the software will likely never be able to simulate the broad diversity of musical manuscript in its rendering engine.
We therefore see most future potential in combining its manuscript images with unsupervised domain adaptation techniques \cite{kang2020unsupervised}, either diffusion models \cite{benigmim2023one,Peng2024unsupervised}, or moment matching methods exploiting batch normalisation, which have seen some success on OMR \cite{Rosell2024SourceFree}.

Despite these limitations, we can still conclude that the high-quality synthetic manuscript images do, in fact, present a significant improvement towards real-world applications of OMR for cultural heritage preservation and access.




%

%
%

\begin{credits}
\subsubsection{\ackname} This work has been supported by the Ministry of Culture of the Czech Republic (project OmniOMR of the NAKI III programme, no. DH23P03OVV008). The computing infrastructure was provided by the LINDAT/CLARIAH-CZ Research Infrastructure,\footnote{\url{https://lindat.cz}} supported by the Ministry of Education, Youth and Sports of the Czech Republic (project no. LM2023062).

\subsubsection{\discintname}
The authors have no competing interests to declare that are
relevant to the content of this article. 
\end{credits}
%
%
%
\bibliographystyle{splncs04}
\bibliography{bibliography}
%




\end{document}